\documentclass[conference]{IEEEtran}
\IEEEoverridecommandlockouts

\usepackage{cite}
\usepackage{amsmath,amssymb,amsfonts}
\usepackage{graphicx}
\usepackage{textcomp}
\usepackage{xcolor}
\usepackage{booktabs}
\usepackage{multirow}
\usepackage{url}
\usepackage[utf8]{inputenc}
\usepackage[T1]{fontenc}
\usepackage{microtype}
\usepackage[font=footnotesize,labelfont=bf]{caption}
\usepackage{subcaption}
\usepackage[none]{hyphenat}
\usepackage{hyperref}
\usepackage{threeparttable}
\usepackage{pifont}
\usepackage{enumitem}
\usepackage[normalem]{ulem}
\usepackage[letterpaper, top=0.75in, bottom=0.75in, left=0.75in, right=0.75in]{geometry}

\newcommand{\cmark}{\ding{51}} 
\newcommand{\xmark}{\ding{55}} 

\begin{document}

\title{\vspace{0.25in}Can VLMs Unlock Semantic Anomaly Detection? A Framework for Structured Reasoning}

\author{\IEEEauthorblockN{Roberto Brusnicki, David Pop, Yuan Gao, Mattia Piccinini, Johannes Betz}
\IEEEauthorblockA{\textit{Professorship of Autonomous Vehicle Systems} \\
\textit{TUM School of Engineering and Design, Technical University of Munich}\\
Munich, Germany}
}

\maketitle

\begin{abstract}
    Autonomous driving systems remain critically vulnerable to the long-tail of rare, out-of-distribution semantic anomalies. While VLMs have emerged as promising tools for perception, their application in anomaly detection remains largely restricted to prompting proprietary models - limiting reliability, reproducibility, and deployment feasibility. To address this gap, we introduce SAVANT (Semantic Anomaly Verification/Analysis Toolkit), a novel model-agnostic reasoning framework that reformulates anomaly detection as a layered semantic consistency verification. By applying SAVANT's two-phase pipeline -- structured scene description extraction and multi-modal evaluation -- existing VLMs improve their scores in detecting anomalous driving scenarios from input images. Our approach replaces ad hoc prompting with semantic-aware reasoning, transforming VLM-based detection into a principled decomposition across four semantic domains. We show that across a balanced set of real-world driving scenarios, applying SAVANT improves VLM's absolute recall by approximately 18.5\% compared to prompting baselines. Moreover, this gain enables reliable large-scale annotation: leveraging the best proprietary model within our framework, we automatically labeled around 10,000 real-world images with high confidence. We use the resulting high-quality dataset to fine-tune a 7B open-source model (Qwen2.5-VL) to perform single-shot anomaly detection, achieving 90.8\% recall and 93.8\% accuracy--surpassing all models evaluated while enabling local deployment at near-zero cost. By coupling structured semantic reasoning with scalable data curation, we provide a practical solution to data scarcity in semantic anomaly detection for autonomous systems. Supplementary material: \texttt{https://TUM-AVS.github.io/SAVANT/}.
\end{abstract}


\section{INTRODUCTION}\label{sec:intro}

The widespread and safe deployment of autonomous vehicles (AVs) depends on their ability to respond well to the rare, low-probability events that are impossible to fully capture in training datasets \cite{yang2024generalized}. Modern autonomous driving systems (ADS) work successfully in the usual ordinary cases but struggle in unexpected scenarios, hindering public trust. Real world failures, such as mistaking a full moon for a traffic light or a billboard stop sign for a real one, highlights this problem. 
These instances reflect critical failures of contextualization that compromise safe adoption of AVs.

\begin{figure}[thpb]
\centering
{\color{white}\framebox{\includegraphics[width=0.97\columnwidth]{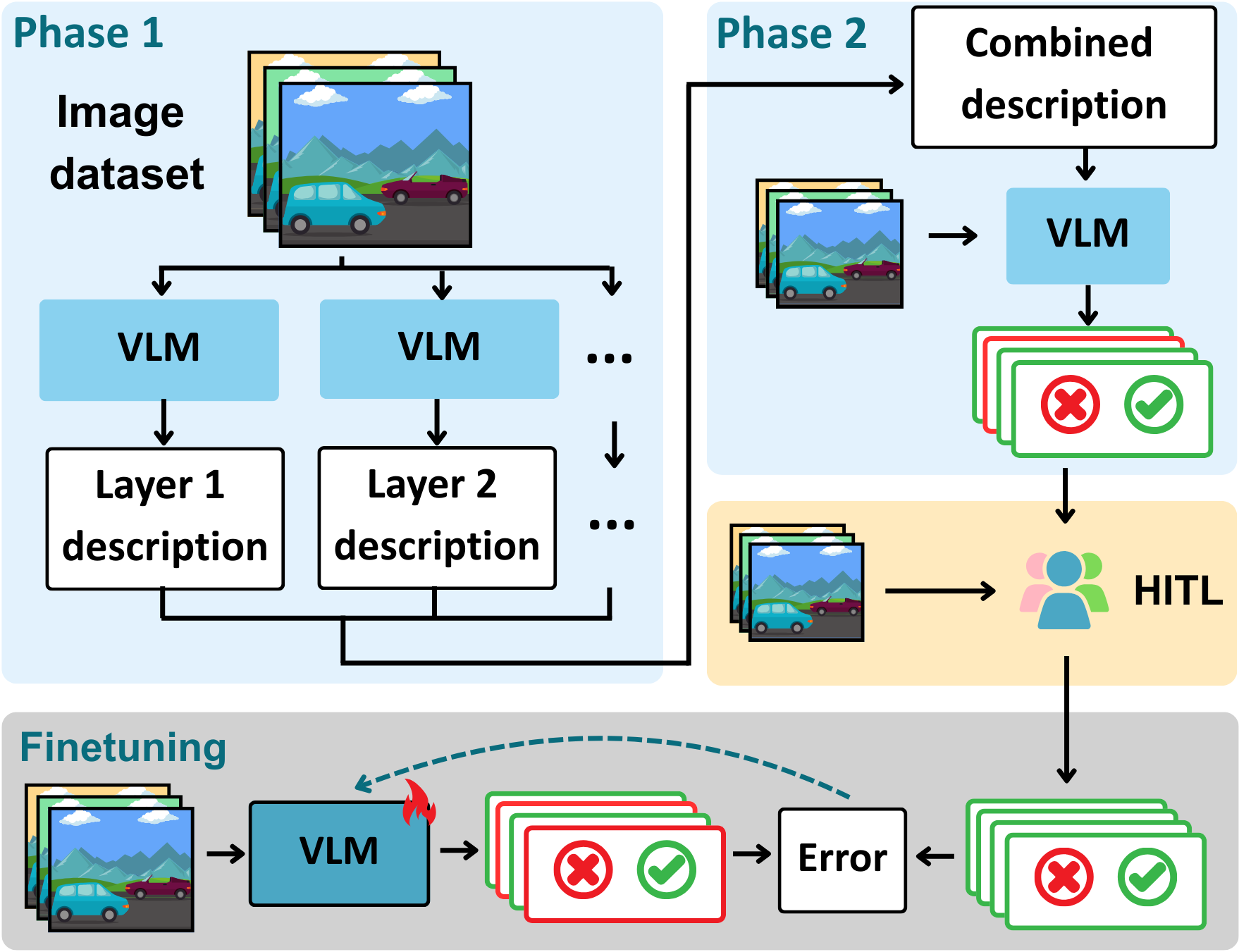}}}
\caption{SAVANT data-model refinement cycle. The final architecture--selected through extensive ablations--operates in two structured phases: (1) Scene Decomposition into Semantic Domains (street, infrastructure, movable objects, environment) and (2) Joint Anomaly Evaluation conditioned on both the image and aggregated layer descriptions. This structured reasoning substantially improves reliability over unstructured prompting, enabling large-scale, high-confidence annotation of driving images. Predictions undergo lightweight Human-in-the-Loop (HITL) verification to efficiently curate high-quality labels, which are then used to fine-tune a compact VLM for single-shot anomaly detection (bottom). The improved model can re-enter the pipeline, establishing a scalable cycle that progressively enhances performance while minimizing human effort.}
\label{fig:savant_overview}
\vspace{-1em}
\end{figure}

\begin{figure}[t]
\centering
{\color{white}\framebox{
\begin{tabular}{@{}c@{\hspace{0.1em}}c@{}}
\includegraphics[width=0.21\textwidth,height=0.21\textwidth]{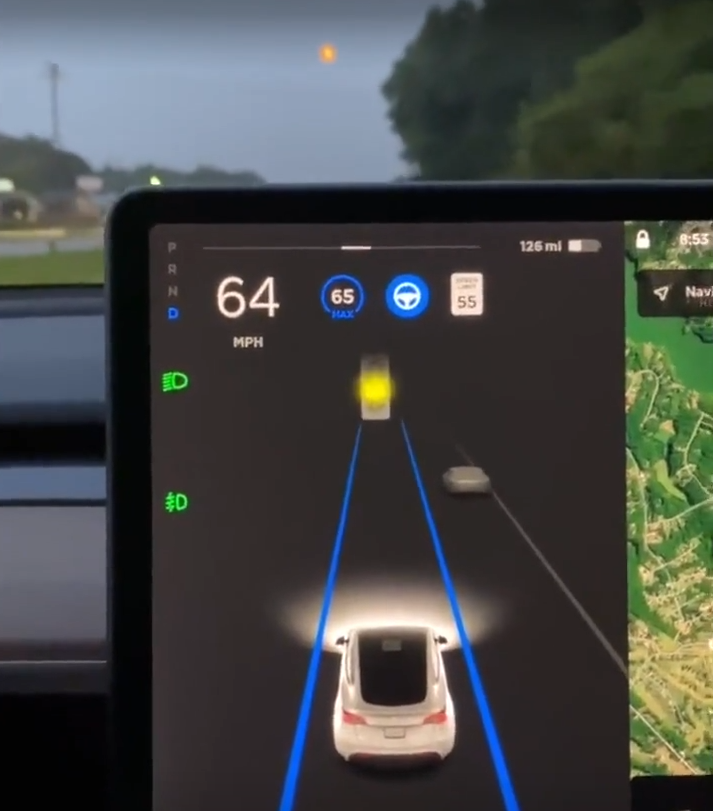} &
\includegraphics[width=0.21\textwidth,height=0.21\textwidth]{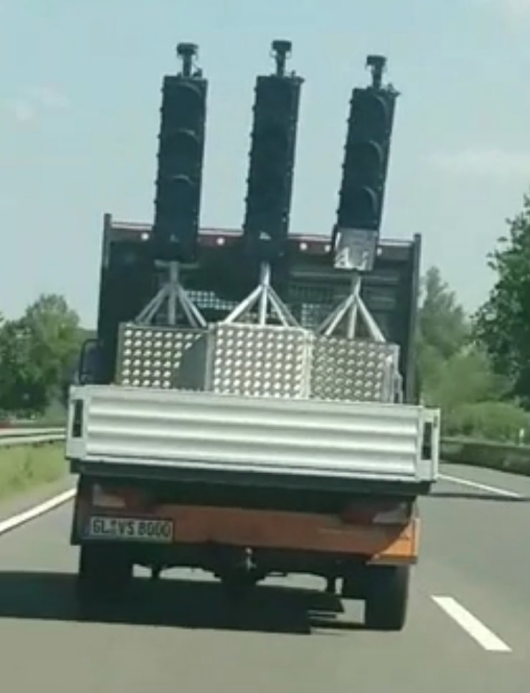} \\
\footnotesize \color{black}(a) Moon as traffic light & \footnotesize \color{black}(b) Traffic lights on truck \\[0.1cm]
\includegraphics[width=0.21\textwidth,height=0.13\textwidth]{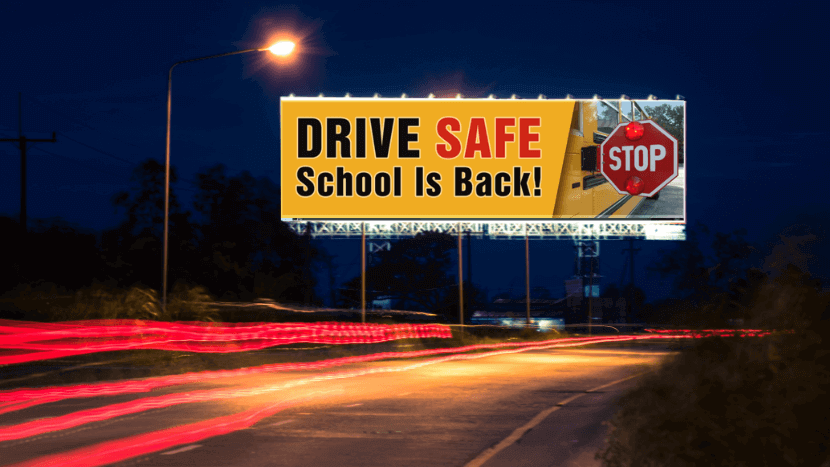} &
\includegraphics[width=0.21\textwidth,height=0.13\textwidth]{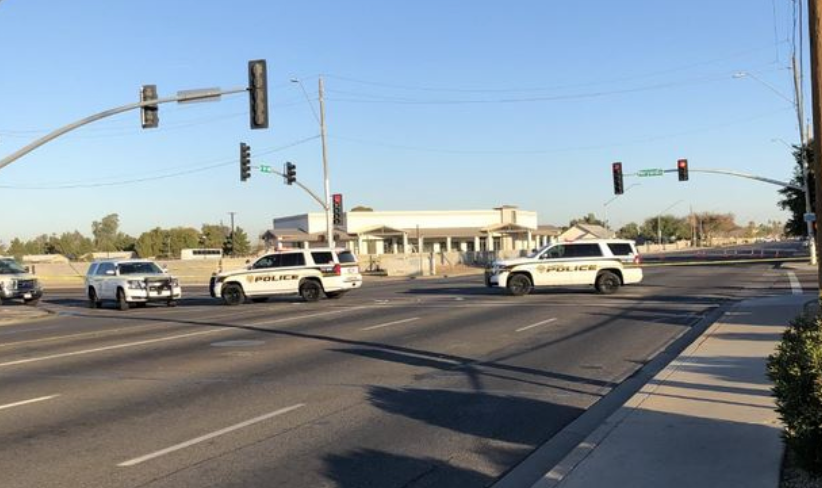} \\
\footnotesize \color{black}(c) Stop sign on billboard & \footnotesize \color{black}(d) Police cars crossing diagonally \\
\end{tabular}
}}
\caption{Real-world semantic anomalies in driving scenarios: individually recognizable objects in contextually unusual configurations that create potentially unsafe or confusing scenarios for autonomous systems.}
\label{fig:ood_examples}
\vspace{-1.5em}
\end{figure}


Vision Language Models (VLMs) leverage vast pretraining on multimodal datasets to bridge the gap between raw perception and semantic reasoning \cite{yang2024llm4drive, zhou2024vision}. These models can interpret complex scenes and reason about spatial configurations that traditional perception systems often miss. While Large Language Models (LLMs) have shown potential as foundation models for autonomous driving \cite{LLM4AD_survey}, recent advancements in VLMs specifically combine visual understanding with natural language reasoning. This makes them well suited for detecting the subtle contextual inconsistencies that characterize long-tail driving anomalies. 

Despite this promise, VLMs face a key limitation for safety-critical anomaly detection: prompting is unreliable and often requires costly proprietary models. Queries frequently lack robustness, interpretability, and consistency required for deployment. Our findings show that prompt-based strategies often miss critical anomalies and produce false positives, while the high latency and cost of large proprietary models make them impractical for real-time use. To overcome these challenges, we propose SAVANT (Semantic Anomaly Verification/Analysis Toolkit), a structured reasoning framework outlined in Figure~\ref{fig:savant_overview}.

\subsection{Related Work}

\subsubsection{Foundation Models for Driving Intelligence}

Recent works use foundation models for autonomous driving in two ways: as end-to-end driving agents and as scene-understanding systems \cite{gao2025foundationmodelsautonomousdriving}. The first trend removes modularity, mapping sensor data directly to vehicle controls. LMDrive, DriveGPT4, EMMA, and ImagiDrive exemplify VLM use in closed-loop driving \cite{shao2023lmdrive, xu2023drivegpt4, emma2024, imagidrive2025}. Hybrid approaches, leverage LLM reasoning while integrating classical control to guide vehicle behavior \cite{sha2023languagempc}. Other architectures, such as DriveVLM and LMAD, couple VLM reasoning with classical perception stacks, highlighting the need for both deliberative and reactive control \cite{tian2024drivevlm, lmad2025, wang2024dualadduallayerplanningreasoning}. While powerful, these end-to-end models are black boxes, limiting safety verification and interpretability, and thus unsuitable for safety-critical monitoring \cite{physical_risk_robotics}.

A different approach uses VLMs for offline scene interpretation. DriveLM reformulates driving as a visual question answering task enabling multistep reasoning, such as analyzing object interactions and their future states \cite{sima2023drivelm}. Benchmarks like CODA-LM and recent pedestrian behavior analyses push VLM scene understanding further \cite{chen2024automated, vlm_pedestrian_2025}. End-to-end systems, though rich in output, are unsuitable for real-time safety monitoring, while Visual Question Answering (VQA) systems remain too detached from control tasks. We fill this gap using VLM reasoning for online safety monitoring.

\subsubsection{Semantic Anomaly Detection}

A key safety requirement for autonomous systems is monitoring inputs and reacting to deviations from training distributions. Beyond machine learning (ML), this includes covariate shifts (e.g., noise, heavy rain) and contextual plausibility of novel objects and scenes \cite{yang2024generalized}. Our focus is on contextual shifts, requiring world understanding beyond statistical pattern matching.

Prior works used object detectors to extract a bag of objects, then applied text-only LLMs to assess plausibility from object co-occurrence \cite{elhafsi2023semantic, sinha2024realtime}. These methods were limited by text-only input (ignoring rich visual data) and by validation in simplified simulations. Other works took data-centric approaches, using VLMs to mine corner cases or generative models to create challenging scenarios \cite{hu2025vlmc4l, hu2023gaia, wang2023drivedreamer, generating_scenes_2025}. These improve training robustness with diverse data, while our method provides a runtime safety net for anomaly detection at inference time.

Hence, existing methods do not allow interpretability, rely on text-only reasoning, or depend on simulation. We instead use multimodal VLMs directly on real-world images, bridging modal and simulation gaps. Our layered method, to our knowledge the first for automated semantic anomaly detection on real driving data, provides direct visual evidence for more reliable and trustworthy autonomous systems.

\subsection{Critical Summary}

The emergence of foundation models is transforming autonomous driving research, moving from modular pipelines to end-to-end systems \cite{yang2024llm4drive, zhou2024vision, LLM4AD_survey}. This advances reasoning and scene understanding, but raises reliability challenges in rare long-tail events. We address this using a VLM to monitor semantic failures—scenes where object arrangements are contextually unsafe (Figure~\ref{fig:ood_examples}) \cite{elhafsi2023semantic, sinha2024realtime}. Building on our preliminary work, we present a structured framework for this task, illustrated in Figure~\ref{fig:savant_overview}.

Current approaches have several key limitations: unstructured prompting lacks robustness, interpretability, and consistency; end-to-end models are black boxes unsuitable for safety-critical monitoring; prior methods rely on text-only reasoning, ignoring visual data; and existing solutions are validated in simulations rather than real-world data. These collectively prevent reliable VLM-based anomaly detection in safety-critical driving applications.

\subsection{Contributions}
\thispagestyle{empty}

To address the previous limitations, focusing in robust anomaly detection and practical deployment for real-world operation, the contributions of this paper are the following:

1) We propose \textbf{SAVANT}, a structured reasoning framework that reformulates anomaly detection as layered semantic consistency verification, improving VLM absolute recall by 18.5\% over prompting baselines.

2) We provide a \textbf{systematic evaluation} analysis of 33 state-of-the-art VLMs across performance, cost, and efficiency dimensions, providing insights for practical deployment of semantic anomaly detection.

3) We demonstrate \textbf{accessible deployment} through fine-tuned 7B open-source models reaching 90.8\% recall and 93.8\% accuracy, surpassing all other models while enabling local deployment without API dependencies.

4) We release \textbf{research resources} including complete framework implementation, optimized prompts for all evaluated models, fine-tuned models, web interface for efficient label correction, and extended CODA dataset with 9,640 annotated real-world driving images to facilitate further research in semantic anomaly detection.

\section{THE SAVANT FRAMEWORK}
\label{sec:methodology}

This section presents SAVANT, our structured reasoning framework for semantic anomaly detection in autonomous driving. We address the limitations of naive VLM prompting through a two-phase approach that decomposes complex driving scenes into analyzable semantic components.

\subsection{Layered Anomaly Formulation}

We formalize semantic anomaly detection as a binary classification task that transcends traditional object-level analysis. Given an input image $\mathit{Img}$ from an autonomous vehicle's forward-facing camera, our goal is to determine whether the scene contains contextually inappropriate arrangements of objects that could compromise system safety.

Unlike conventional out-of-distribution detection that identifies unknown objects, semantic anomalies involve familiar elements in contextually invalid configurations. Let $\mathcal{O} = \{o_1, o_2, \ldots, o_n\}$ denote the set of objects detected in a scene, and let $\mathcal{C}$ represent the expected context--the set of spatial, relational, and functional constraints that govern valid object configurations in a driving environment. We define a \emph{semantic anomaly} as a scene where $\mathcal{O}$ violates $\mathcal{C}$:
\begin{equation}
A(\mathit{Img}) = \begin{cases} 1, & \text{if } \exists \; (o_i, o_j) \in \mathcal{O}^2 : R(o_i, o_j) \notin \mathcal{C} \\ 0, & \text{otherwise} \end{cases}
\label{eq:anomaly_def}
\end{equation}
where $A(\mathit{Img}) = 1$ indicates a semantic anomaly. $R(o_i, o_j)$ captures the spatial, functional, or contextual relationship between objects $o_i$ and $o_j$. Crucially, each $o_i \in \mathcal{O}$ may be individually common in driving scenes; the anomaly arises from their \emph{combination} or \emph{context}. For example, traffic lights being transported on a truck represent a semantic anomaly: both ``traffic light'' and ``truck'' are common driving objects, but their relationship $R(\text{traffic\_light}, \text{truck}) = \text{carried\_on}$ violates the expected context $\mathcal{C}$, creating a potentially dangerous scenario.

Our approach decomposes scene-level anomaly detection into four semantic layers, each capturing distinct aspects of traffic scenes linked to anomalies:

\textbf{Layer 1: Street (S)} - Road topology, geometry, surface conditions, and lane markings that define the driving surface and its structural integrity.

\textbf{Layer 2: Infrastructure (I)} - Devices, signs, signals, and barriers that regulate traffic flow and provide guidance.

\textbf{Layer 3: Movable Objects (M)} - Vehicles, pedestrians, and dynamic entities that navigate through the environment.

\textbf{Layer 4: Environment (E)} - Weather, lighting, and visibility conditions that affect scene perception and safety.

These four layers are adapted from the 6-Layer Model for urban traffic description \cite{scholtes2021sixlayer}, excluding two layers: digital information, absent from current anomaly datasets, and temporal modifications, where VLMs showed poor temporal understanding in prior experiments. Extending to these layers is left for future work. The resulting structure provides broad scene coverage with minimal redundancy, enabling systematic analysis while maintaining interpretability and targeted anomaly identification per semantic domain.

\subsection{Two-Phase Anomaly Detection Pipeline}

SAVANT employs a structured two-phase pipeline that transforms unstructured VLM analysis into systematic reasoning, addressing the limitations of naive prompting.

\textbf{Phase 1: Structured Scene Description Extraction.} Rather than directly querying for anomaly detection, we first guide the VLM to systematically describe the scene according to our four-layer decomposition. For each semantic layer $l \in \{1,2,3,4\}$, we employ carefully designed prompt templates $P_l$ that direct the model's attention to specific aspects of the scene image  $\mathit{Img}$:
\vspace{-0.4em}
\begin{equation}
D_l = \mathrm{VLM}(\mathit{Img}, P_l)
\end{equation}
where $D_l$ represents the textual description extracted for layer $l$. This structured extraction forces the model to examine each semantic aspect systematically, reducing the likelihood of overlooking critical details that might indicate anomalies. The complete scene description aggregates information across all layers:
\vspace{-0.4em}
\begin{equation}
D_{scene} = \mathrm{Aggregate}(D_1, D_2, D_3, D_4)
\end{equation}

This phase provides interpretable intermediate representations, and creates rich textual context for the subsequent evaluation phase.

\textbf{Phase 2: Multi-Modal Scene Evaluation.} The second phase leverages both visual and textual information for robust anomaly classification. The VLM receives the original image and the structured scene description $D_{scene}$ as joint inputs:
\vspace{-0.4em}
\begin{equation}
\mathrm{Classification} = \mathrm{VLM}(\mathit{Img}, D_{scene}, P_{eval})
\end{equation}
where $P_{eval}$ is the evaluation prompt that instructs the model to analyze the scene for semantic inconsistencies using both visual evidence and the extracted textual descriptions. This combines visual perception with the structured reasoning from textual analysis. The evaluation follows a systematic process: layer-wise anomaly assessment, cross-layer interaction analysis, and final binary classification with supporting rationale. Our results will show how this structured approach improves both accuracy and interpretability compared to direct prompting methods.

\subsection{Fine-tuning Integration Strategy}

While our two-phase pipeline achieves high accuracy, multiple VLM queries limit real-time deployment. To address this, we leverage SAVANT's high-quality outputs as an automated data annotation engine.

We apply our structured framework to automatically label large-scale datasets, generating training data that captures the reasoning embedded in SAVANT's two-phase approach. Using this data, we fine-tune compact VLMs to internalize the structured reasoning:
\vspace{-0.4em}
\begin{equation}
f_{fine\text{-}tuned}(\mathit{Img}) = \mathrm{VLM}_{fine\text{-}tuned}(\mathit{Img}, P_{direct})
\end{equation}
where $P_{direct}$ is a single evaluation prompt that instructs the model to perform anomaly classification directly from the image, without explicit intermediate layer decomposition (Fig.~\ref{fig:savant_overview}). This strategy enables us to distill SAVANT's multi-phase reasoning into an efficient single-shot model suitable for real-time deployment. The fine-tuned model maintains the benefits of structured analysis while achieving the computational efficiency required for practical autonomous system integration. Crucially, this approach enables smaller, open-source models to achieve performance levels that rival or exceed larger proprietary alternatives, providing a practical and accessible path toward widespread deployment.

\section{EXPERIMENTAL SETUP}
\label{sec:experimental_setup}

This section describes our evaluation of SAVANT across three axes: performance against baselines, scalability across VLM architectures, and practical deployment.

\subsection{Datasets}

We build three incremental datasets, as described below.

\textbf{Model Selection.} We begin evaluation with \textbf{CODALM\_small}, a dataset that we curate to contain 100 real-world driving images (50 anomalous, 50 normal) derived from the CODA corner case dataset \cite{li2022codarealworldroadcorner}. Each image receives manual annotation with detailed textual scene descriptions and anomaly evaluations, ensuring a high-quality ground truth. This dataset enables our initial scanning across 30+ VLM candidates: the small size of the dataset allows us to identify the best models of each family without excessive resource consumption, while providing sufficient data for measuring average response times and API costs.

\begin{figure}[!t]
\centering
\includegraphics[width=0.95\columnwidth]{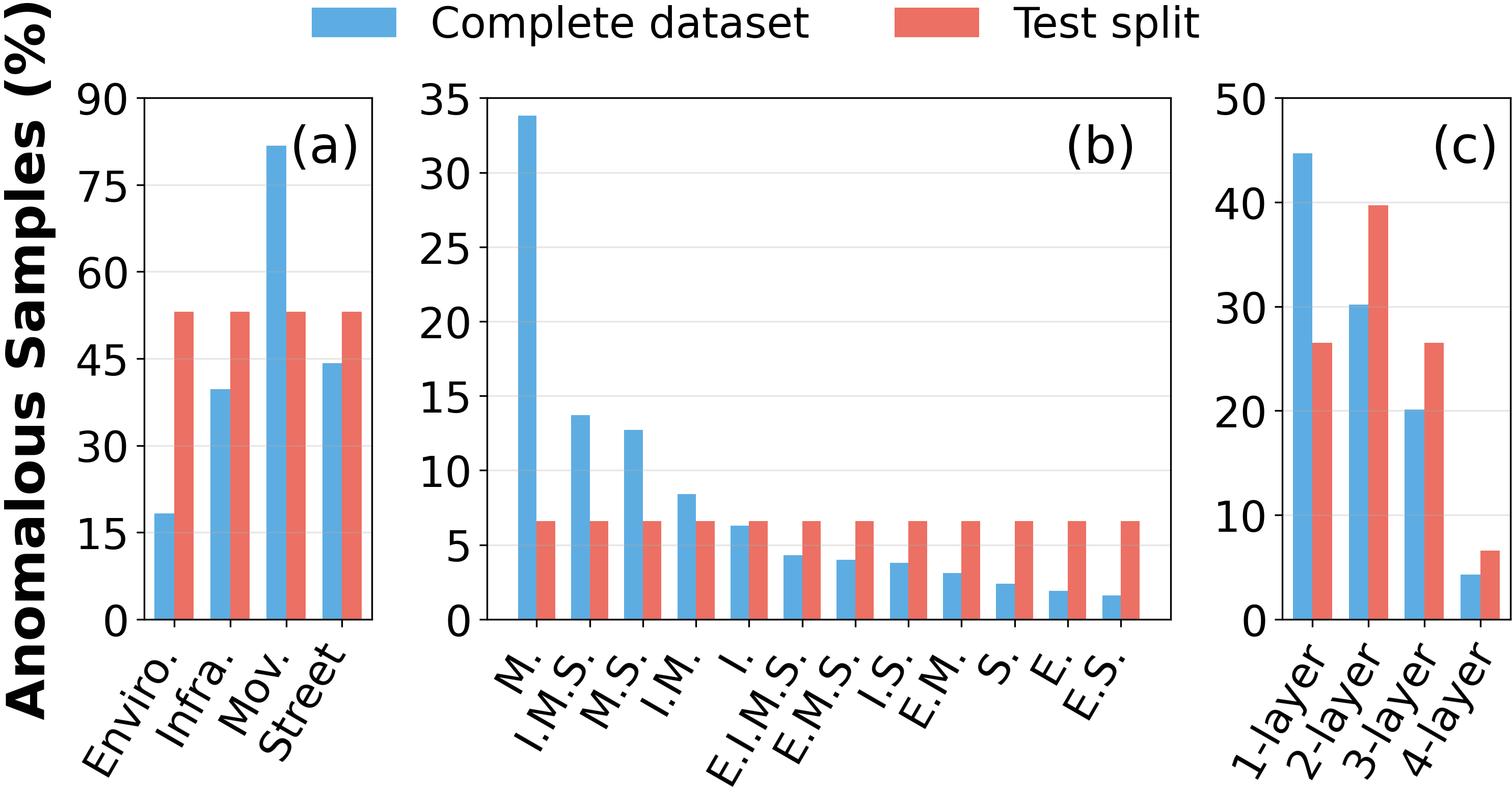}
\caption{Layer-wise anomaly distribution comparing CODALM\_medium dataset (5,078 samples) and its test split (1020 samples):(a) Individual layer frequency across the four semantic layers, (b) Anomaly layer combination frequency, (c) Multi-layer anomaly frequency distribution.}
\label{fig:codalm_medium_distribution}
\vspace{-1em}
\end{figure}

\textbf{Comparative Evaluation.} To validate our approach, we created \textbf{CODALM\_medium} by combining automated framework evaluation with human validation. Starting with the full CODA dataset (9,640 images), we used Gemini-2.0-Flash-Exp (our top-performing model at that time) to generate scene descriptions and anomaly evaluations. From these, two human evaluators reviewed and corrected 5,078 annotations, producing a balanced dataset with validated ground truth. Of these, 1,020 form the balanced test subset; the rest serve as fine-tuning data. Figure~\ref{fig:codalm_medium_distribution} shows the anomaly distribution across CODALM\_medium: 60.9\% contain anomalies spanning four semantic layers. Movable Objects anomalies are most frequent (81.7\%), followed by Street Layer (44.2\%), Infrastructure (39.7\%), and Environmental (18.3\%). The dataset captures varying complexity, from single-layer (27.4\%) to quad-layer anomalies (2.6\%), providing broad coverage for training robust models.

\textbf{Framework Application.} Finally, we release the complete 9,640-image \textbf{CODALM\_large} dataset, fully labeled using SAVANT with the best-performing VLM. This dataset represents the largest semantically-annotated dataset for anomaly detection in autonomous driving, and demonstrates our framework's capability as a scalable data annotation engine.

\subsection{Models Evaluated}

Our evaluation encompasses both proprietary and open-source VLMs to provide baseline comparisons and assess the accessibility of different approaches.

\textbf{Proprietary Models.} We evaluate the best available models from major commercial VLM families accessed via API, including representatives from Google Gemini, OpenAI GPT, Anthropic Claude, Mistral, and Qwen-VL families. These models represent the current pinnacle of VLM capabilities but require ongoing costs and external dependencies.

\textbf{Open-source Models.} We include leading open-source alternatives from key model families, including Qwen2.5-VL variants, Mistral models, Pixtral, Gemma, and LLaVA architectures spanning different parameter scales from 7B to 72B. This comparison shows the broad applicability of SAVANT across diverse model architectures.

\subsection{Evaluation Methods}

We evaluate multiple configurations, grouped into three baselines (1--3) without layered reasoning, and four structured methods (4--7) using our framework:

\begin{enumerate}[leftmargin=*, itemsep=0.3em]
    \item \texttt{image\_baseline}: Direct VLM prompting with images only, representing the naive approach of asking: ``Is this traffic scene anomalous? Yes or No.'' This baseline captures unstructured, single-shot VLM performance without our layered reasoning framework.
    \item \texttt{text\_baseline}: Uses unstructured descriptions for evaluation, without visual grounding or layered analysis.
    \item \texttt{baseline}: Combines unstructured scene description with image-based evaluation.
    \item \texttt{image}: Direct image analysis using structured four-layer decomposition without textual scene descriptions.
    \item \texttt{text}: Uses structured descriptions from Phase 1 with layered evaluation, without visual grounding in Phase 2.
    \item \texttt{full}: Complete SAVANT pipeline combining structured scene descriptions with multi-modal layered evaluation.
    \item \texttt{*\_opt}: Optimized versions using DSPy \cite{khattab2024dspy} MIPROv2 prompt optimization for enhanced performance.
\end{enumerate}

\begin{table}[t]
\vspace*{0.5em}
\caption{Evaluation configurations and their properties.}
\label{tab:evaluation_configurations}
\centering
\begin{tabular}{l@{\hspace{1.5em}}l@{\hspace{1.5em}}c@{\hspace{1em}}c@{\hspace{1em}}c}
\toprule
        \textbf{Method} & \textbf{Input} & \textbf{L} & \textbf{O} & \textbf{\#Q} \\
\midrule
        \texttt{image\_baseline} & image & \xmark & \xmark & 1 \\
        \texttt{text\_baseline} & scene description & \xmark & \xmark & 2 \\
        \texttt{baseline} & image + scene description & \xmark & \xmark & 2 \\
        \midrule
        \texttt{image} & image & \cmark & \xmark & 1 \\
        \texttt{text} & scene description & \cmark & \xmark & 5 \\
        \texttt{text\_opt} & scene description & \cmark & \cmark & 5 \\
        \texttt{full} & image + scene description & \cmark & \xmark & 5 \\
        \texttt{full\_opt} & image + scene description & \cmark & \cmark & 5 \\
\bottomrule
\end{tabular}\\[0.5em]
\footnotesize{L = Layered analysis, O = Optimization, \#Q = Number of queries}
\vspace{-1em}
\end{table}

Table~\ref{tab:evaluation_configurations} summarizes the key properties of each evaluation method, including input modalities, structural components, and computational requirements.

These setups enable systematic analysis of our framework's core innovations: structured reasoning, multimodality, layered decomposition, and prompt optimization effects.

\subsection{Evaluation Metrics and Implementation}

\textbf{Performance Metrics.} We report precision, recall, F1-score, and accuracy across all experiments. For safety-critical anomaly detection, recall is paramount as missing true anomalies poses greater risk than false alarms.

\textbf{Image Resolution.} All models process images at 360p resolution, representing an optimal trade-off between performance and computational efficiency based on our comprehensive resolution analysis presented in Section~\ref{sec:results}. 

\textbf{Efficiency Metrics.} We measure inference time (average seconds per query) and cost analysis (average input/output tokens per query) to assess practical deployment feasibility.

\textbf{Implementation Details.} Experiments utilize DSPy \cite{khattab2022demonstrate} for prompt optimization, applying MIPROv2 optimization with few-shot examples where applicable. All inferences use temperature 0 for deterministic, reproducible results.

\textbf{Prompt Structure.} In Phase 1, prompts instruct the VLM to describe the specific semantic layer of the scene. Phase 2's  prompt provides the image alongside the concatenated layer descriptions and asks the model to classify the scene as anomalous or normal, with a supporting rationale. Exact prompt templates are provided in our public repository.

\textbf{Fine-tuning.} We employ Low-Rank Adaptation for parameter-efficient supervised fine-tuning of Qwen2.5-VL-7B-Instruct, with a standard cross-entropy loss over the target classification tokens. Training is conducted for 3 epochs with a learning rate of $2 \times 10^{-4}$ and batch size of 4.

\textbf{Data Filtering.} Generated annotations from the pipeline are filtered by a two-stage process: (1) automated consistency checks reject samples where the model's classification contradicts its own rationale or where outputs are malformed; (2) human evaluators review and correct a subset of remaining annotations through our web-based curation interface, resolving ambiguous cases and correcting systematic errors.

\section{RESULTS AND ANALYSIS}
\label{sec:results}

SAVANT's effectiveness is demonstrated experimentally: structured reasoning and DSPy optimization improve VLM anomaly detection  enabling accessible deployment.

\subsection{Baseline Performance: Image-Only Evaluations}

Table~\ref{tab:comprehensive_vlm_comparison} shows baseline results for 32 state-of-the-art VLMs using the \texttt{image\_baseline} method—direct image-only prompting without structured reasoning. Models receive only visual input with a simple prompt instructing it to classify the traffic scene. Response times $T$ are average inference times, and token counts show total tokens (input + output) per query. This establishes a performance floor and identifies the top-models per family for subsequent structured reasoning evaluation.

This evaluation reveals clear performance hierarchies: top proprietary models (Gemini 2.5 Pro: 85\%, GPT-5: 83\%) outperform open-source alternatives (Qwen2.5-VL 72B: 75\%, Mistral Small 3.1: 74\%), with inference times varying from 2.6 to 24.9 seconds.

\begin{table}[t]
\vspace*{0.5em}
\caption{VLMs performance for image-only anomaly detection}
\label{tab:comprehensive_vlm_comparison}
\centering
\begin{tabular}{l@{\hspace{0.7em}}|@{\hspace{0.7em}}c@{\hspace{0.7em}}c@{\hspace{0.7em}}c@{\hspace{0.7em}}c@{\hspace{0.7em}}|@{\hspace{0.7em}}c@{\hspace{0.7em}}c}
\toprule
        \textbf{Proprietary Models} & \textbf{Accu.} & \textbf{Prec.} & \textbf{Rec.} & \textbf{F1} & \textbf{T(s)} & \textbf{Tokens} \\
\midrule
        Gemini 2.5 Pro & \textbf{0.85} & 0.94 & 0.70 & \underline{0.80} & 24.9 & 938 \\
        GPT-5 & \underline{0.83} & 0.88 & \underline{0.77} & \textbf{0.82} & 28.7 & 51851 \\
        Gemini 2.5 Flash & 0.81 & 0.91 & 0.69 & 0.78 & 10.6 & 947 \\
        Gemini 1.5 Flash & 0.81 & 0.90 & 0.70 & 0.79 & 2.9 & 958 \\
        Mistral Medium 3.1 & 0.80 & 0.80 & \textbf{0.80} & 0.80 & 7.9 & 1259 \\
        Mistral Medium 3 & 0.81 & 0.94 & 0.66 & 0.78 & 6.4 & 1158 \\
        Qwen-VL Max & 0.79 & 0.87 & 0.68 & 0.76 & 7.7 & 2193 \\
        GPT-4o & 0.78 & 0.94 & 0.60 & 0.73 & 7.4 & 1032 \\
        Gemini 2.5 Pro Prev & 0.77 & 0.87 & 0.64 & 0.74 & 23.6 & \textbf{898} \\
        Claude 3.5 Sonnet & 0.74 & 0.79 & 0.66 & 0.72 & 7.3 & 1109 \\
        Gemini 2.0 Flash Exp & 0.73 & 0.80 & 0.62 & 0.70 & \textbf{2.6} & 2441 \\
        Gemini 2.0 Thinking & 0.75 & 0.90 & 0.56 & 0.69 & 8.1 & \underline{937} \\
        Gemini 2.5 Flash Prev & 0.75 & 0.90 & 0.56 & 0.69 & 11.2 & 956 \\
        GPT-4 Turbo & 0.73 & \textbf{0.96} & 0.48 & 0.64 & 7.6 & 1055 \\
        Claude 3.5 Haiku & 0.72 & 0.89 & 0.50 & 0.64 & 7.0 & 1144 \\
        Claude Sonnet 4 & 0.70 & 0.83 & 0.50 & 0.63 & 9.8 & 1145 \\
        Claude Opus 4.1 & 0.68 & \underline{0.95} & 0.38 & 0.54 & 19.4 & 1238 \\
        Claude 3.7 Sonnet & 0.65 & 0.80 & 0.40 & 0.53 & 10.8 & 1276 \\
        GPT-4.1 Mini & 0.66 & 0.94 & 0.34 & 0.50 & 5.4 & 1075 \\
        GPT-4o Mini & 0.65 & 0.90 & 0.34 & 0.49 & 4.4 & 14787 \\
        Claude Opus 4 & 0.62 & 0.83 & 0.31 & 0.45 & 13.1 & 1224 \\
        Qwen-VL Plus & 0.58 & 0.90 & 0.18 & 0.30 & \underline{2.8} & 2155 \\
        GPT-4.1 Nano & 0.56 & 0.80 & 0.16 & 0.27 & 4.3 & 1262 \\
\midrule
\textbf{Open Models} & \textbf{Accu.} & \textbf{Prec.} & \textbf{Rec.} & \textbf{F1} & \textbf{T(s)} & \textbf{Tokens} \\
\midrule
        Qwen2.5-VL 72B & \textbf{0.75} & 0.82 & 0.64 & \underline{0.72} & 9.4 & \underline{961} \\
        Mistral Small 3.1 & \underline{0.74} & 0.71 & \textbf{0.82} & \textbf{0.76} & 11.9 & 1057 \\
        Pixtral Large 2411 & 0.67 & 0.64 & \underline{0.78} & 0.70 & 5.8 & 1817 \\
        Mistral Small 3.2 & 0.70 & 0.78 & 0.56 & 0.65 & 5.5 & 996 \\
        Qwen2.5-VL 32B & 0.62 & 0.65 & 0.53 & 0.59 & 19.9 & 2380 \\
        Gemma3 12B & 0.44 & 0.46 & 0.70 & 0.56 & 9.2 & 1254 \\
        Pixtral 12B & 0.64 & \textbf{1.00} & 0.28 & 0.44 & 5.4 & 1637 \\
        LLaVA 1.5 7B & 0.58 & 0.79 & 0.22 & 0.34 & 6.7 & 1029 \\
        Qwen2.5-VL 7B & 0.55 & 0.55 & 0.48 & 0.52 & \underline{4.2} & \textbf{686} \\
        Qwen2.5-VL 3B & 0.50 & 0.50 & 0.29 & 0.37 & \textbf{3.6} & 1489 \\
\bottomrule
\end{tabular}
\vspace{-1em}
\end{table}

\begin{table*}[t!]
\vspace*{0.5em}
\caption{Performance comparison of different methods across representative VLMs at 360p resolution.}
\label{tab:main_results}
\centering
\footnotesize
\begin{tabular}{l@{}|@{}c@{}c@{}c@{}c@{}|@{}c@{}c@{}c@{}c@{}|@{}c@{}c@{}c@{}c@{}|@{}c@{}c@{}c@{}c@{}|@{}c@{}c@{}c@{}c}
\multicolumn{1}{l|}{ } & \multicolumn{4}{c|}{\textbf{Gemini-2.0-FE}} & \multicolumn{4}{c|}{\textbf{Qwen2.5-VL-72B}} & \multicolumn{4}{c|}{\textbf{Qwen2.5-VL-32B}} & \multicolumn{4}{c|}{\textbf{Qwen2.5-VL-7B}} & \multicolumn{4}{c}{\textbf{Qwen2.5-VL-3B}} \\
\toprule
\textbf{Method} & \textbf{\phantom{X}Acc.} & \textbf{\phantom{X}Rec.} & \textbf{\phantom{X}Prec} & \textbf{\phantom{X}F1\phantom{X}} & \textbf{\phantom{X}Acc.} & \textbf{\phantom{X}Rec.} & \textbf{\phantom{X}Prec} & \textbf{\phantom{X}F1\phantom{X}} & \textbf{\phantom{X}Acc.} & \textbf{\phantom{X}Rec.} & \textbf{\phantom{X}Prec} & \textbf{\phantom{X}F1\phantom{X}} & \textbf{\phantom{X}Acc.} & \textbf{\phantom{X}Rec.} & \textbf{\phantom{X}Prec} & \textbf{\phantom{X}F1\phantom{X}} & \textbf{\phantom{X}Acc.} & \textbf{\phantom{X}Rec.} & \textbf{\phantom{X}Prec} & \textbf{\phantom{X}F1\phantom{X}} \\
\midrule
image\_baseline{\color{white}-} & 0.73 & 0.62 & 0.79 & 0.70 & 0.75 & 0.64 & 0.82 & 0.72 & 0.62 & 0.53 & \uline{0.65} & 0.59 & 0.55 & 0.48 & 0.55 & 0.52 & 0.50 & 0.29 & 0.50 & 0.37 \\
text\_baseline & 0.65 & 0.40 & 0.80 & 0.53 & 0.69 & 0.48 & \textbf{0.85} & 0.61 & 0.61 & 0.43 & \textbf{0.67} & 0.53 & 0.59 & 0.39 & \textbf{0.65} & 0.49 & 0.50 & 0.35 & 0.50 & 0.41 \\
baseline & 0.77 & 0.66 & \uline{0.85} & 0.74 & 0.73 & 0.56 & \uline{0.83} & 0.67 & \uline{0.65} & 0.72 & 0.63 & 0.67 & 0.60 & 0.46 & 0.64 & 0.53 & \uline{0.56} & \textbf{0.47} & \uline{0.57} & \uline{0.52} \\
\midrule
image & 0.78 & \textbf{0.94} & 0.71 & 0.81 & 0.69 & \textbf{0.86} & 0.64 & 0.74 & 0.64 & 0.62 & 0.65 & 0.63 & 0.45 & 0.36 & 0.43 & 0.39 & 0.46 & 0.28 & 0.44 & 0.34 \\
text & 0.77 & \uline{0.92} & 0.71 & 0.80 & 0.79 & \uline{0.84} & 0.76 & \uline{0.80} & 0.60 & 0.48 & 0.62 & 0.54 & 0.43 & 0.31 & 0.40 & 0.35 & 0.51 & 0.22 & 0.52 & 0.31 \\
text\_opt & 0.80 & 0.86 & 0.77 & 0.81 & \textbf{0.82} & 0.82 & 0.82 & \textbf{0.82} & 0.64 & 0.71 & 0.62 & 0.66 & \textbf{0.61} & \uline{0.53} & \uline{0.64} & \uline{0.58} & 0.51 & 0.45 & 0.51 & 0.48 \\
\textbf{full} & \uline{0.85} & 0.90 & 0.82 & \uline{0.86} & \uline{0.80} & 0.78 & 0.81 & \uline{0.80} & 0.64 & \uline{0.74} & 0.62 & \uline{0.67} & 0.59 & 0.53 & 0.60 & 0.56 & 0.45 & 0.30 & 0.43 & 0.35 \\
\textbf{full\_opt} & \textbf{0.88} & 0.90 & \textbf{0.87} & \textbf{0.88} & \textbf{0.82} & \uline{0.84} & 0.81 & \textbf{0.82} & \textbf{0.66} & \textbf{0.75} & 0.63 & \textbf{0.69} & \uline{0.60} & \textbf{0.65} & 0.64 & \textbf{0.62} & \textbf{0.59} & \uline{0.45} & \textbf{0.63} & \textbf{0.52} \\
\bottomrule
\end{tabular}
\vspace{-1.5em}
\end{table*}

\begin{figure}[!t]
\centering
{\color{white}\framebox{\includegraphics[width=0.97\columnwidth]{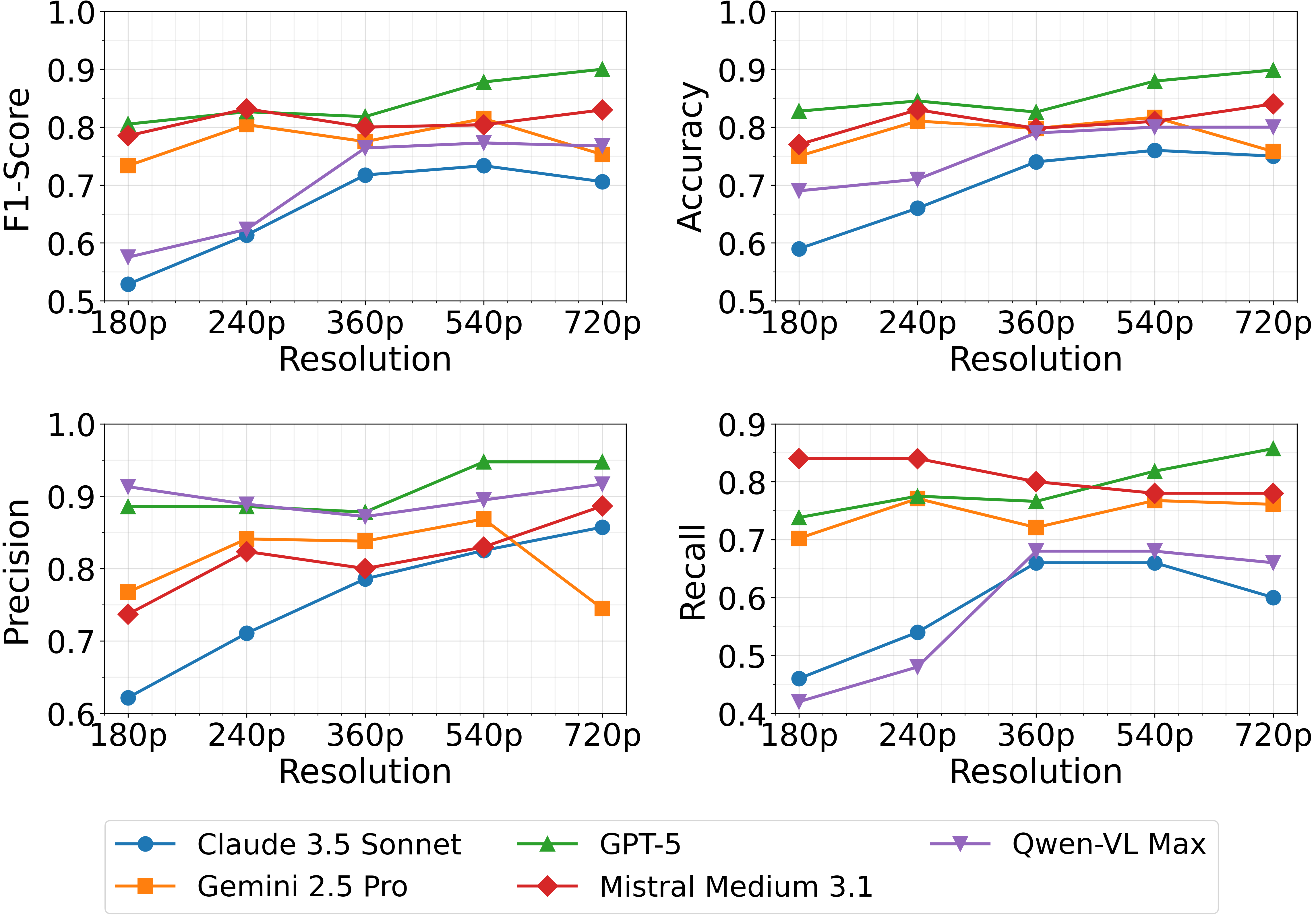}}}
\caption{Resolution analysis for top models of each family. Performance improves up to 360p with diminishing returns at higher resolutions.}
\label{fig:resolution_comparison}
\vspace{-1em}
\end{figure}

Figure~\ref{fig:resolution_comparison} shows performance scores (F1, accuracy, precision, recall) for top models per family evaluated on a balanced split of 1000 examples across resolutions of 180p, 240p, 360p, 540p, and 720p. Performance improvements up to 360p for most models, with only marginal gains from 360p to 540p or 720p. Considering the substantial increase in token consumption costs for higher resolutions (2.25x for 540p, 4x for 720p), 360p represents the optimal balance between performance and efficiency.

\subsection{Structured Reasoning Outperforms Naive Prompting}

Our first key finding demonstrates that SAVANT's layered analysis approach significantly outperforms unstructured baseline methods across all evaluated VLMs. Table~\ref{tab:main_results} presents our core performance comparison using Gemini-2.0-Flash-Exp, selected over the higher-performing Gemini-2.5-Pro due to cost considerations

Our \texttt{image\_baseline} serves as the critical single-prompt control: it queries the VLM with only ``Is this traffic scene anomalous? Yes or No,'' without any structured decomposition or layered reasoning. This isolates the contribution of SAVANT's four-layer framework from the inherent capabilities of the underlying VLM. When paired with Gemini-2.0-Flash-Exp, SAVANT's full framework achieves 90\% recall and 85\% accuracy, representing a 36\% relative improvement in recall over the two-phase baseline (absolute improvement of 24\%) and a 45\% relative improvement over the single-prompt \texttt{image\_baseline}. Most notably, the DSPy-optimized version reaches 88\% accuracy while maintaining 90\% recall, demonstrating that structured reasoning enables reliable anomaly detection even in safety-critical scenarios where missing true positives carries significant risk.

The bigger improvement in text-only scenarios (40\% to 92\% recall) highlights the importance of structured descriptions. Our approach captures critical semantic information that unstructured descriptions miss, proving that systematic decomposition enhances VLM reasoning capabilities.

However, our extensive evaluation across the Qwen2.5-VL family reveals a persistent performance gap between proprietary and open-source models. Despite extensive experimentation with multiple model sizes (from 3B  up to 72B parameters) and evaluation methods, the best-performing open model (Qwen2.5-VL-72B with 82\% F1) falls short of the proprietary baseline (Gemini-2.0-Flash-Exp with 88\% F1). This gap motivated our investigation into fine-tuning approaches to achieve competitive performance with locally deployable, cost-effective solutions.

\subsection{The Critical Importance of Multi-Modal Evaluation}

Our second key finding establishes that combining visual and textual information in SAVANT's Phase 2 evaluation provides superior performance compared to text-only reasoning. This validates our design choice of multi-modal evaluation rather than purely textual analysis after the description extraction in Phase 1.

In Table \ref{tab:main_results}, our layered approach shows substantial improvements in image-only evaluation for high-capacity models, where recall increases from 62\% to 94\% for Gemini-2.0-Flash-Exp when VLMs are guided to analyze specific semantic layers before classification. This demonstrates that explicit reasoning guidance enhances VLM performance on complex visual scenes for models with sufficient representational capacity, though the effect varies across model architectures.

Our multi-modal approach leverages both the richness of visual perception and the structured reasoning provided by textual analysis, consistently outperforming single-modality alternatives across all evaluated configurations.

\subsection{Model Scale and Optimization Effectiveness}

Table~\ref{tab:main_results} reveals critical insights about the relationship between model scale and optimization effectiveness. DSPy optimization shows varying benefits across model sizes: while the 72B model benefits from optimization (text: 80\% $\rightarrow$ text\_opt: 82\% F1), smaller models show mixed results. The 7B model demonstrates substantial optimization gains (text: 35\% $\rightarrow$ text\_opt: 58\% F1), while the 3B model shows moderate improvement (text: 31\% $\rightarrow$ text\_opt: 48\% F1), indicating that optimization effectiveness depends on both model capacity and the specific optimization target.

This scaling relationship has profound implications for deployment strategies. Larger models (72B, 32B) can leverage sophisticated reasoning frameworks effectively, while smaller models (7B, 3B) require alternative approaches such as fine-tuning to achieve competitive performance. The consistent performance degradation from 72B (82\% F1) $\rightarrow$ 32B (69\% F1) $\rightarrow$ 7B (62\% F1) $\rightarrow$ 3B (52\% F1) establishes clear trade-offs between model accessibility and task performance.

Furthermore, structured reasoning methods (image, text, full) generally outperform their baseline counterparts across model architectures, demonstrating that explicit decomposition of the reasoning process benefits VLMs. However, the magnitude and consistency of improvement varies significantly with model capacity and architecture, reinforcing the importance of model selection for deployment scenarios with varying computational constraints.

\subsection{Fine-Tuned Open Model Outperforms Proprietary Ones}

Our key contribution addresses the deployment challenge through fine-tuning. Table~\ref{tab:finetuning_results} shows our main result: SAVANT enables a fine-tuned 7B open-source model to rival proprietary models while enabling local deployment.

\begin{table}[t]
\vspace*{0.5em}
\caption{Fine-Tuned vs. Baseline vs. Top models Performance}
\label{tab:finetuning_results}
\centering
\begin{tabular}{l@{\hspace{0.5em}}|@{\hspace{0.5em}}c@{\hspace{0.5em}}c@{\hspace{0.5em}}c@{\hspace{0.5em}}c@{\hspace{0.5em}}|@{\hspace{0.5em}}c@{\hspace{0.5em}}c@{\hspace{0.5em}}|@{\hspace{0.5em}}c}
\toprule
        \textbf{Model} & \textbf{Acc.} & \textbf{Rec.} & \textbf{Prec.} & \textbf{F1} & \textbf{T(s)} & \textbf{Tok.} & \textbf{Q} \\
\midrule
        Qwen2.5-VL-7B (NFT) & 0.55 & 0.48 & 0.55 & 0.52 & \uline{1.74} & \uline{105} & \textbf{1} \\
        Qwen2.5-VL-7B (FT) & \textbf{0.94} & \textbf{0.91} & \textbf{0.97} & \textbf{0.94} & \textbf{0.08} & \textbf{2} & \textbf{1} \\
        Qwen2.5-VL-7B (PFT) & 0.84 & 0.82 & 0.85 & 0.83 & 13.85 & 788 & \uline{2} \\
        Gemini-2.0-FE (full\_opt) & \uline{0.88} & \uline{0.90} & \uline{0.86} & \uline{0.88} & 17.8 & 791 & 5 \\
\bottomrule
\end{tabular}\\[0.5em]
\footnotesize{T = Inference time, Tok. = Output tokens, Q = Number of queries}
\vspace{-2em}
\end{table}

The fine-tuning yields two complementary solutions. The single-shot model classify scenes with a single query without intermediate steps. The pipeline model keeps SAVANT's two-phase structure (description extraction - multi-modal evaluation) using fine-tuned components. The single-shot model achieves 93.8\% accuracy and 90.8\% recall, while the pipeline model reaches 83.7\% accuracy and 81.8\% recall. Both boost recall from the 48.4\% baseline while maintaining computational efficiency.

Notably, the single-shot model outperforms the pipeline variant despite the latter’s use of structured reasoning that proved beneficial in our prior evaluation. This counterintuitive result stems from the increased training complexity of the pipeline, which must learn both description generation and multimodal anomaly detection simultaneously. Both models were trained with identical hyperparameters and epochs, but the pipeline model’s additional complexity may require extended training to reach its full potential, suggesting that even higher scores could be achieved.

The expanded comparison in Table~\ref{tab:finetuning_results} demonstrates that our single-shot fine-tuned model (93.8\% accuracy) surpasses all proprietary baselines including Gemini 2.5 Pro (85\% accuracy), GPT-4o (78\% accuracy), and Claude 3.5 Sonnet (74\% accuracy). Our pipeline fine-tuned model (83.7\% accuracy) also outperforms GPT-4o and Claude 3.5 Sonnet while requiring only two queries compared to the five queries needed by Gemini-2.0-FE's full optimized approach.

This comparison shows that our fine-tuned models substantially outperform their baseline version and achieve competitive results that match or exceed proprietary alternatives. The performance transformation from baseline (54.6\% accuracy, 48.4\% recall) to fine-tuned variants (84-94\% accuracy, 82-91\% recall) demonstrates a significant improvement in accessibility for practical anomaly detection deployment. Computational cost analysis is provided in the supplementary material.

\begin{figure}[ht]
\centering
\includegraphics[width=0.9\columnwidth]{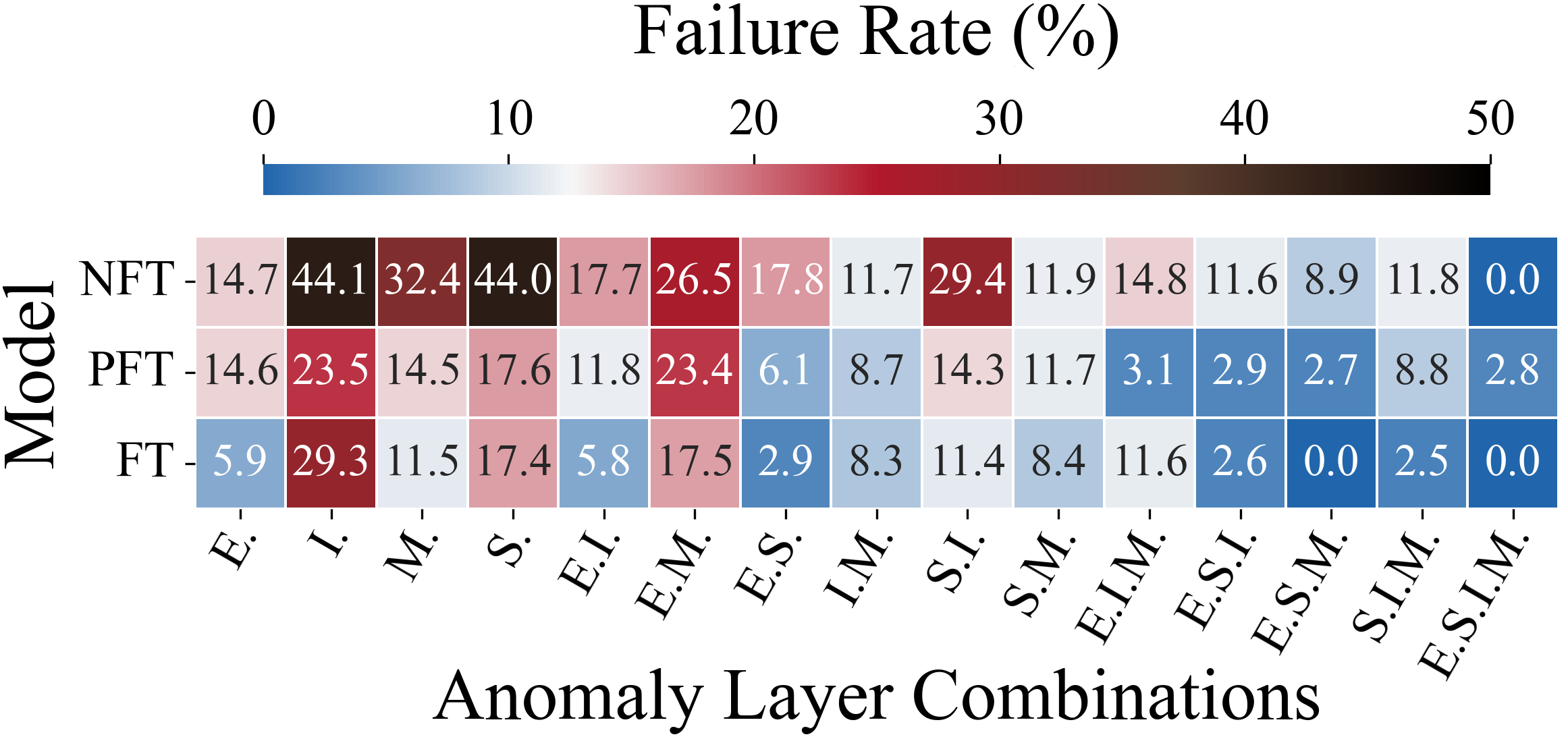}
\caption{Failure rates (\%) across semantic layer combinations for three Qwen2.5-VL-7B variants: FT (Fine-Tuned), PFT (Pipeline Fine-Tuned), and NFT (Non-Fine-Tuned). Layer abbreviations: S (Street), I (Infrastructure), M (Movable Objects), E (Environmental). Multi-layer combinations (e.g., E.S.I.M) represent anomalies spanning multiple semantic contexts.}
\label{fig:failure_rates_heatmap}
\vspace{-1em}
\end{figure}

\subsection{Layer-Specific Error Analysis}

To understand how fine-tuning affects error patterns across SAVANT's semantic decomposition, we analyze layer-wise failure rates for the three Qwen2.5-VL-7B variants from Table~\ref{tab:finetuning_results}. Figure~\ref{fig:failure_rates_heatmap} compares the fine-tuned single-shot (FT), pipeline fine-tuned (PFT), and non-fine-tuned baseline (NFT) models across all anomaly layer combinations.

The FT model exhibits the lowest failure rates, remaining below 10\% for most anomalies, particularly across Street (S) and Environment (E) layer combinations. This demonstrates that fine-tuning effectively internalizes SAVANT's structured reasoning. In contrast, the NFT baseline shows high error rates exceeding 30\% for most single-layer anomalies--especially Infrastructure (I) and Movable Objects (M)--confirming that without fine-tuning, the 7B model struggles with semantic layer associations.

The PFT variant achieves moderate gains over baseline but exhibits elevated failure rates ($\approx$ 40\%) for Infrastructure-related anomalies (I, I.M, S.I.), supporting our observation that two-phase reasoning complexity may hinder training convergence.

Across all variants, Environmental (E) anomalies remain most difficult due to their subtle nature (e.g., lighting, fog). Interestingly, failure rates decrease for multi-layer scenarios (e.g., E.S.I.M), where cross-layer cues improve detection through redundant semantic evidence.

This analysis reinforces two key findings: (1) single-shot fine-tuning enables robust, context-aware detection across all semantic layers, and (2) Environmental conditions remain the primary challenge--requiring future work on visual-semantic reasoning under adverse conditions. Qualitative failure examples and layer-wise outputs are provided in the supplementary material.

\section{CONCLUSION}

In this paper, we addressed the critical challenge of semantic anomaly detection in autonomous driving, where data scarcity and unreliable VLM performance have hindered practical deployment. We introduced SAVANT, a structured reasoning framework that transforms ad-hoc VLM prompting into systematic analysis across four semantic layers: Street, Infrastructure, Movable Objects, and Environment. Through comprehensive evaluation of 33 state-of-the-art VLMs, we demonstrated that structured reasoning significantly outperforms unstructured baselines, with SAVANT achieving 89.6\% recall and 88.0\% accuracy.

Our framework addresses multiple critical gaps simultaneously: it enables researchers to efficiently evaluate large numbers of models, provides systematic prompt optimization through DSPy integration, and facilitates efficient human curation of model-generated annotations. This multi-faceted approach produces high-quality labeled data that enables fine-tuning open-source models to achieve state-of-the-art performance at dramatically reduced costs. Most significantly, our fine-tuned 7B Qwen2.5-VL model achieves 90.8\% recall and 93.8\% accuracy, surpassing all evaluated proprietary models while enabling cost-free local deployment.

By automatically annotating 9,640 real-world driving images and demonstrating scalable data curation workflows, SAVANT offers a practical approach to address data scarcity in semantic anomaly detection. This work demonstrates the potential for accessible, reliable safety monitoring in autonomous driving. For future work, we plan to extend SAVANT to temporal analysis through video input and validate the framework through real-world on-vehicle integration.


\begin{thebibliography}{99}

\bibitem{yang2024generalized} J. Yang, K. Zhou, Y. Li, and Z. Liu, "Generalized Out-of-Distribution Detection: A Survey," International Journal of Computer Vision, vol. 132, no. 12, pp. 5635--5662, 2024, doi: 10.1007/s11263-024-02117-4.

\bibitem{yang2024llm4drive} Z. Yang, X. Jia, H. Li, and J. Yan, "LLM4Drive: A Survey of Large Language Models for Autonomous Driving," arXiv preprint arXiv:2311.01043, 2024.

\bibitem{zhou2024vision} X. Zhou, M. Liu, E. Yurtsever, B. L. Zagar, W. Zimmer, H. Cao, and A. C. Knoll, "Vision Language Models in Autonomous Driving: A Survey and Outlook," arXiv preprint arXiv:2310.14414, 2024.

\bibitem{LLM4AD_survey} C. Cui, Y. Ma, S.-Y. Park, Z. Yang, Y. Zhou, J. Lu, J. Peng, J. Zhang, R. Zhang, L. Li, Y. Chen, J. H. Panchal, A. Abdelraouf, R. Gupta, K. Han, and Z. Wang, "LLM4AD: Large Language Models for Autonomous Driving -- Concept, Review, Benchmark, Experiments, and Future Trends," arXiv preprint arXiv:2410.15281, 2025.

\bibitem{gao2025foundationmodelsautonomousdriving} Y. Gao, M. Piccinini, Y. Zhang, D. Wang, K. Moller, R. Brusnicki, B. Zarrouki, A. Gambi, J. F. Totz, K. Storms, S. Peters, A. Stocco, B. Alrifaee, M. Pavone, and J. Betz, "Foundation Models in Autonomous Driving: A Survey on Scenario Generation and Scenario Analysis," IEEE Open Journal of Intelligent Transportation Systems, pp. 1--1, 2026, doi: 10.1109/OJITS.2026.3660686.

\bibitem{shao2023lmdrive} H. Shao, Y. Hu, L. Wang, G. Song, S. L. Waslander, Y. Liu, and H. Li, "LMDrive: Closed-Loop End-to-End Driving with Large Language Models," in 2024 IEEE/CVF Conference on Computer Vision and Pattern Recognition (CVPR), 2024, pp. 15120--15130, doi: 10.1109/CVPR52733.2024.01432.

\bibitem{xu2023drivegpt4} Z. Xu, Y. Zhang, E. Xie, Z. Zhao, Y. Guo, K.-Y. K. Wong, Z. Li, and H. Zhao, "DriveGPT4: Interpretable End-to-End Autonomous Driving Via Large Language Model," IEEE Robotics and Automation Letters, vol. 9, no. 10, pp. 8186--8193, 2024, doi: 10.1109/LRA.2024.3440097.

\bibitem{emma2024} J.-J. Hwang, R. Xu, H. Lin, W.-C. Hung, J. Ji, K. Choi, D. Huang, T. He, P. Covington, B. Sapp, Y. Zhou, J. Guo, D. Anguelov, and M. Tan, "EMMA: End-to-End Multimodal Model for Autonomous Driving," Transactions on Machine Learning Research, 2025.

\bibitem{imagidrive2025} J. Li, B. Zhang, X. Jin, J. Deng, X. Zhu, and L. Zhang, "ImagiDrive: A Unified Imagination-and-Planning Framework for Autonomous Driving," arXiv preprint arXiv:2508.11428, 2025.

\bibitem{sha2023languagempc} H. Sha, Y. Mu, Y. Jiang, L. Chen, C. Xu, P. Luo, S. E. Li, M. Tomizuka, W. Zhan, and M. Ding, "LanguageMPC: Large Language Models as Decision Makers for Autonomous Driving," arXiv preprint arXiv:2310.03026, 2025.

\bibitem{tian2024drivevlm} X. Tian, J. Gu, B. Li, Y. Liu, Y. Wang, Z. Zhao, K. Zhan, P. Jia, X. Lang, and H. Zhao, "DriveVLM: The Convergence of Autonomous Driving and Large Vision-Language Models," in Conference on Robot Learning (CoRL), 2024.

\bibitem{lmad2025} N. Song, B. Zhang, X. Zhu, J. Deng, and L. Zhang, "LMAD: Integrated End-to-End Vision-Language Model for Explainable Autonomous Driving," arXiv preprint, 2025.

\bibitem{wang2024dualadduallayerplanningreasoning} D. Wang, M. Kaufeld, and J. Betz, "DualAD: Dual-Layer Planning for Reasoning in Autonomous Driving," in 2025 IEEE/RSJ International Conference on Intelligent Robots and Systems (IROS), 2025, pp. 12057--12063, doi: 10.1109/IROS60139.2025.11246084.

\bibitem{physical_risk_robotics} T. Kojima, Y. Zhu, Y. Iwasawa, T. Kitamura, G. Yan, S. Morikuni, R. Takanami, A. Solano, T. Matsushima, A. Murakami, and Y. Matsuo, "A Comprehensive Survey on Physical Risk Control in the Era of Foundation Model-enabled Robotics," in Proceedings of the Thirty-Fourth International Joint Conference on Artificial Intelligence (IJCAI-25), Survey Track, 2025.

\bibitem{sima2023drivelm} C. Sima, K. Renz, K. Chitta, L. Chen, H. Zhang, C. Xie, J. Bei{\ss}wenger, P. Luo, A. Geiger, and H. Li, "DriveLM: Driving with Graph Visual Question Answering," in European Conference on Computer Vision (ECCV), Oral, 2024.


\bibitem{chen2024automated} K. Chen, Y. Li, W. Zhang, Y. Liu, P. Li, R. Gao, L. Hong, M. Tian, X. Zhao, Z. Li, D.-Y. Yeung, H. Lu, and X. Jia, "Automated Evaluation of Large Vision-Language Models on Self-Driving Corner Cases," in 2025 IEEE/CVF Winter Conference on Applications of Computer Vision (WACV), 2025, pp. 7817--7826, doi: 10.1109/WACV61041.2025.00759.

\bibitem{vlm_pedestrian_2025} H. Gao, L. Zhang, Y. Zhao, Z. Yang, and J. Cao, "Application of Vision-Language Models to Pedestrian Behavior Prediction and Scene Understanding in Autonomous Driving," in 2025 4th International Conference on Robotics, Artificial Intelligence and Intelligent Control (RAIIC), 2025, pp. 01--08, doi: 10.1109/RAIIC65850.2025.11170191.

\bibitem{elhafsi2023semantic} A. Elhafsi, R. Sinha, C. Agia, B. Schmerling, E. Pavone, and M. Pavone, "Semantic anomaly detection with large language models," Autonomous Robots, vol. 47, pp. 1035--1055, 2023, doi: 10.1007/s10514-023-10132-6.

\bibitem{sinha2024realtime} R. Sinha, A. Elhafsi, C. Agia, M. Foutter, E. Schmerling, and M. Pavone, "Real-Time Anomaly Detection and Reactive Planning with Large Language Models," in Robotics: Science and Systems (RSS), 2024, doi: 10.15607/RSS.2024.XX.114.

\bibitem{hu2025vlmc4l} H. Hu, J. Zuo, Y. Lou, Y. Cui, J. Wang, N. Guan, J. Wang, Y.-H. Li, and C. J. Xue, "VLM-C4L: Continual Core Dataset Learning with Corner Case Optimization via Vision-Language Models for Autonomous Driving," arXiv preprint arXiv:2503.23046, 2025.

\bibitem{hu2023gaia} A. Hu, L. Russell, H. Yeo, Z. Murez, G. Fedoseev, A. Kendall, J. Shotton, and G. Corrado, "GAIA-1: A Generative World Model for Autonomous Driving," arXiv preprint arXiv:2309.17080, 2023.

\bibitem{wang2023drivedreamer} X. Wang, Z. Zhu, G. Huang, X. Chen, J. Zhu, and J. Lu, "DriveDreamer: Towards Real-World-Drive World Models for Autonomous Driving," in Computer Vision -- ECCV 2024, 2025, pp. 55--72.


\bibitem{generating_scenes_2025} Y. Wu, H. Zhang, T. Lin, L. Huang, S. Luo, R. Wu, C. Qiu, W. Ke, and T. Zhang, "Generating Multimodal Driving Scenes via Next-Scene Prediction," in 2025 IEEE/CVF Conference on Computer Vision and Pattern Recognition (CVPR), 2025, pp. 6844--6853, doi: 10.1109/CVPR52734.2025.00642.

\bibitem{scholtes2021sixlayer} M. Scholtes, L. Westhofen, L. R. Turner, K. Lotto, M. Schuldes, H. Weber, N. Wagener, C. Neurohr, M. H. Bollmann, F. K\"{o}rtke, J. Hiller, M. Hoss, J. Bock, and L. Eckstein, "6-Layer Model for a Structured Description and Categorization of Urban Traffic and Environment," IEEE Access, vol. 9, pp. 59131--59147, 2021, doi: 10.1109/ACCESS.2021.3072739.

\bibitem{li2022codarealworldroadcorner} K. Li, K. Chen, H. Wang, L. Hong, C. Ye, J. Han, Y. Chen, W. Zhang, C. Xu, D.-Y. Yeung, X. Liang, Z. Li, and H. Xu, "CODA: A Real-World Road Corner Case Dataset for Object Detection in Autonomous Driving," in Computer Vision -- ECCV 2022, 2022, pp. 406--423, doi: 10.1007/978-3-031-19839-7\_24.

\bibitem{khattab2024dspy} O. Khattab, A. Singhvi, P. Maheshwari, Z. Zhang, K. Santhanam, S. Vardhamanan, S. Haq, A. Sharma, T. T. Joshi, H. Moazam, H. Miller, M. Zaharia, and C. Potts, "DSPy: Compiling Declarative Language Model Calls into Self-Improving Pipelines," in International Conference on Learning Representations (ICLR), 2024.

\bibitem{khattab2022demonstrate} O. Khattab, K. Santhanam, X. L. Li, D. Hall, P. Liang, C. Potts, and M. Zaharia, "Demonstrate-Search-Predict: Composing retrieval and language models for knowledge-intensive NLP," arXiv preprint arXiv:2212.14024, 2023.

\end{thebibliography}
\end{document}